\documentclass[10pt, a4paper]{article}

\usepackage[]{lrec-coling2024} 

\usepackage{amsmath}
\usepackage{booktabs}
\usepackage{adjustbox}
\usepackage{amsfonts}
\usepackage{comment}

\title{Schroedinger's Threshold: When the AUC doesn't predict Accuracy}

\name{Juri Opitz} 

\address{University of Zurich \\
         opitz.sci@gmail.com\\}

\abstract{
The Area Under Curve measure (AUC) seems apt to evaluate and compare diverse models, possibly without calibration. An important example of AUC application is the evaluation and benchmarking of models that predict faithfulness of generated text. But we show that the AUC yields an academic and optimistic notion of accuracy that can misalign with the actual accuracy observed in application, yielding significant changes in benchmark rankings. To paint a more realistic picture of downstream model performance (and prepare a model for actual application), we explore different calibration modes, testing calibration data and method. 
 \\ \newline \Keywords{Classification evaluation, AUC score, accuracy, calibration, faithfulness evaluation}
}
\begin{document}

\maketitleabstract

\section{Introcuction}

\begin{figure}
    \centering
    \includegraphics[width=\linewidth, trim=0 0.0cm 1.6cm 0]{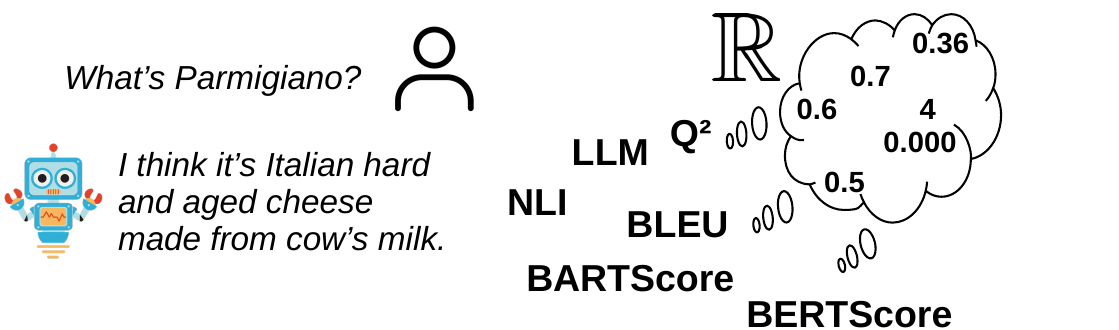}
    \caption{In NLP we witness diverse domains and tasks (here: dialog, faithfulness), and wonder about the predictive power of scores by diverse models (here: e.g., the BERT/BARTscore metric, task-focused systems such as the automatic Q/A metric `Q$^2$' or Natural Language Inference systems, possibly also LLMs). While the AUC seems appealing as an assessment measure, it bears pitfalls.}
    \label{fig:diversity}
\end{figure}

In Natural Language Processing (NLP), we often want to compare diverse models in diverse domains and tasks. Consider Figure \ref{fig:diversity} that shows the answer of a dialog system to a user input. On the machine-generated output, we would like to use a model to judge whether the answer is faithful.\footnote{This particular task is well motivated: Today, text generation models produce millions of texts each day, and their output can still be unfaithful, with some assessing that LLM hallucination are inevitable \cite{xu2024hallucination}. Thus, models that can reliably and efficiently assess faithfulness of generated text are of growing importance \citep{falke-etal-2019-ranking, kryscinski-etal-2020-evaluating, wang-etal-2020-asking, maynez-etal-2020-faithfulness,gekhman2023trueteacher,zha2023alignscore, steen-etal-2023-little, eacl-faith}.} For this, we could draw from a huge shelf of models, including in/out-domain trained classifiers, or even metrics such as BERTscore \cite{bertscore}.

 But how do we evaluate and compare such diverse models? When the target labels are \textit{binary}, e.g., as they are indeed for text faithfulness (but also in many other NLP/ML tasks), it seems appealing to employ the Area Under Curve (AUC) measure. Indeed, AUC has a nice probabilistic interpretation and makes model calibration (i.e., searching for a decision threshold) unnecessary. Mainly for these reasons, the AUC has been explicitly recommended for evaluation and benchmarking of models that predict faithfulness \cite{honovich-etal-2022-true-evaluating,gekhman2023trueteacher,zha2023alignscore}. 

Yet, an issue is that AUC has an academic view on model power. In a ``real-world'' application, we cannot forgo model calibration, as we ultimately have to make decisions. In our example of text faithfulness, there are clear ramifications of different decision thresholds: with a false-positive we run a risk of releasing false or even harmful output; a false-negative may lead to censoring of good system output. 

In this paper, we show that such important real world considerations tend to be neglected by the AUC, and find that its theoretical perspective on system performance may not align with actual performance in applications. Our findings indicate that a main factor for this lies in the diversity of model score and data distributions. We thus argue that AUC should not be used as a sole measure for model evaluation and benchmarking, particularly when models and data are diverse.

In sum, our main contribution is two-fold:

1. We show that the evaluation of diverse models with AUC can be misleading, and that AUC predicts mostly only the optimistic scenario of direct in-domain and in-distribution calibration. 


2. We test different calibration strategies (varying development domain and method) for i) learning how to develop calibrated classifiers from diverse models and ii) best estimate their expected downstream classification performance. 

Our code is available at \url{https://github.com/flipz357/SchroedingersEvaluation}.

\section{Preliminaries}

\begin{figure}
    \centering
    \includegraphics[width=0.7\linewidth, trim=0 0.5cm 0 0]{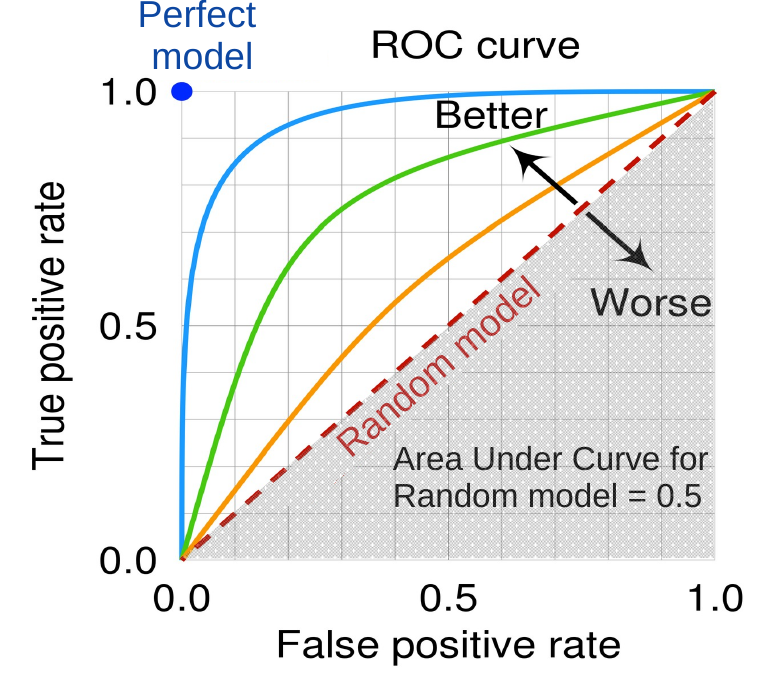}
    \caption{ROC curve examples of different models.}
    \label{fig:rocc}
\end{figure}

\paragraph{AUC (or AUROC)} is the \textit{Area Under the Receiver Operating Characteristic Curve} \cite{fawcett2006introduction}. Given data $\{(x_i, y_i)\}_{i=1}^n$ with $y_i$ a binary label and $x_i$ an input mapped by a model to a score $s_i \in \mathbb{R}$, we can set threshold $\hat{\theta}$ to get a true positive rate $TPR(\hat{\theta})$ and false positive rate $FPR(\hat{\theta})$: 
\begin{equation*}
    TPR(\hat{\theta}) = \frac{TP_{\hat{\theta}}}{TP_{\hat{\theta}} + FN_{\hat{\theta}}} ~~~~~~~~~ FPR(\hat{\theta}) = \frac{FP_{\hat{\theta}}}{FP_{\hat{\theta}} + TN_{\hat{\theta}}}.
\end{equation*}
Given $I[c]$ returns 1 if the condition $c$ is true, and 0 else, the $TP_{\hat{\theta}}$ is the amount of true positives $\sum_{i=1}^{n} I[s_i > \hat{\theta} \land  y_i = 1]$; $TN_{\hat{\theta}}$ is the amount of true negatives $\sum_{i=1}^{n} I[s_i \leq \hat{\theta} \land  y_i = 0]$;  $FP_{\hat{\theta}}$ is the amount of false positives $\sum_{i=1}^{n} I[s_i > \hat{\theta} \land  y_i = 0]$ and $FN_{\hat{\theta}}$ the amount of false negatives $\sum_{i=1}^{n} I[s_i \leq \hat{\theta} \land  y_i = 1]$.  With this, we can plot the receiver-operator curve (ROC) with TPR on the y-axis and FPR on the x-axis, and get the \textit{area under curve} (AUC), which equals 1 for a perfect classifier and 0.5 for a random classifier (cf.\ Figure \ref{fig:rocc}\footnote{Figure under public CC-BY-SA-4.0 license from public domain and further refined by the authors of this paper.}).

The AUC score has an intuitive interpretation: Given two data instances with opposing labels, the AUC score tells us the probability that our model assigns a greater score to the positively labeled instance than to the instance with the negative label.

\paragraph{AUC seems appealing (theoretically):} Besides its intuitive interpretation, the AUC score allows simple evaluation by factoring out model calibration (determining a threshold). Thus we can assess and compare seemingly fairly the \textit{theoretical classification power} of diverse models such as metrics as well as non-calibrated classifiers (e.g., classifiers trained on different domains), and of course also standard classifiers that are already calibrated. 

However, with this theoretic view on model power, the AUC makes us potentially neglect the final goal of most NLP systems: they should assign categorical decisions and show decision skill. If we'd presume that calibration of diverse models would be of same difficulty for any model, relying on AUC would perhaps seem fine. However, diverse models may return diverse score distributions. Data for finding a suitable threshold also can be diverse and noisy. Therefore we hypothesize that calibration suitability of models is also diverse, possibly affecting their real-world classification performance, with ramifications for the utility of AUC.

\section{Experimental setup}

\paragraph{Data sets} are adopted from the popular TRUE benchmark \citep{honovich-etal-2022-true-evaluating}. TRUE combines a rich variety of faithfulness domains in a standardized format: summarization \citep{pagnoni-etal-2021-understanding, maynez-etal-2020-faithfulness, wang-etal-2020-asking, fabbri-etal-2021-summeval}, knowledge-grounded dialog \citep{honovich-etal-2021-q2, gupta-etal-2022-dialfact, begin}, and paraphrases \cite{zhang-etal-2019-paws}.\footnote{Summarization: \textit{quags, summeval, frank, quags-x, quags-c}. Dialog: \textit{begin, dialfact, q2}. Paraphrase: \textit{paws}. 
} TRUE explicitly recommends AUC evaluation.

\textbf{Metrics} that we include are BERTscore \cite{bertscore} using either DeBERTa \cite{he2020deberta}, henceforth denoted by \texttt{DBERTsc}, or RoBERTa \cite{liu2019roberta}, denoted by \texttt{RBERTsc}. As recommended by \citet{honovich-etal-2022-true-evaluating}, we take their \textit{precision} predictions, which should better assess faithfulness than F1 or recall. Then we also show \texttt{BARTsc}(ore) \cite{bartscore}, \texttt{BLEURT} \cite{bleurt} and \texttt{BLEU} (k=4) \cite{papineni-etal-2002-bleu}. 

\textbf{Models} are also diverse. Some are NLI-based (a closely related task), while others employ elaborate scoring techniques, e.g., by analyzing a cross-product of sentences. As in TRUE, we employ \texttt{ANLI} \citep{honovich-etal-2022-true-evaluating} which is a T5-11B \citep{t5-new} LLM trained on ANLI \cite{nie-etal-2020-adversarial}. \texttt{SummacZS}  \citep{laban-etal-2022-summac} evaluates an NLI model on sentence pairs and averages maximum entailment probabilities, and \texttt{Q2} \citep{honovich-etal-2021-q2} integrates a question-answering step.

\subsection{Measurement of expected accuracy}

\begin{table*}[ht!]
    \centering
    \scalebox{0.72}{
    \begin{tabular}{lllllllllllll}
data set & \texttt{BLEU} & \texttt{QuestE} & \texttt{FactCC} & \texttt{SummaCC} & \texttt{SummacZS} & \texttt{BARTSc} & \texttt{RBERTSc} & \texttt{Q2} & \texttt{ANLI} & \texttt{DBERTSc} & \texttt{BLEURT} \\
 \midrule
qags-c & 63.9 | \textbf{11} & 64.2 | \textbf{10} & 76.4 | \textbf{6} & 79.6 | \textbf{5} & 80.9 | \textbf{3} & 80.9 | \textbf{4} & 74.8 | \textbf{7} & 83.5 | \textbf{1} & 82.1 | \textbf{2} & 69.1 | \textbf{9} & 71.6 | \textbf{8} \\
summeval & 60.2 | \textbf{11} & 70.1 | \textbf{9} & 75.9 | \textbf{6} & 79.8 | \textbf{3} & 81.7 | \textbf{1} & 73.5 | \textbf{7} & 73.0 | \textbf{8} & 78.8 | \textbf{4} & 80.5 | \textbf{2} & 77.2 | \textbf{5} & 66.7 | \textbf{10} \\
frank & 78.0 | \textbf{10} & 84.0 | \textbf{7} & 76.4 | \textbf{11} & 88.9 | \textbf{3} & 89.1 | \textbf{2} & 86.1 | \textbf{5} & 80.8 | \textbf{9} & 87.8 | \textbf{4} & 89.4 | \textbf{1} & 84.3 | \textbf{6} & 82.8 | \textbf{8} \\
qags-x & 48.6 | \textbf{11} & 56.3 | \textbf{7} & 64.9 | \textbf{5} & 76.1 | \textbf{3} & 78.1 | \textbf{2} & 53.8 | \textbf{8} & 52.8 | \textbf{9} & 70.9 | \textbf{4} & 83.8 | \textbf{1} & 49.5 | \textbf{10} & 57.2 | \textbf{6} \\
dialfact & 72.5 | \textbf{7} & 77.3 | \textbf{5} & 55.3 | \textbf{11} & 81.2 | \textbf{3} & 84.1 | \textbf{2} & 65.6 | \textbf{8} & 62.9 | \textbf{10} & 86.1 | \textbf{1} & 77.7 | \textbf{4} & 64.2 | \textbf{9} & 73.1 | \textbf{6} \\
mnbm & 49.3 | \textbf{11} & 65.3 | \textbf{6} & 59.4 | \textbf{10} & 67.2 | \textbf{4} & 71.3 | \textbf{2} & 60.9 | \textbf{9} & 65.5 | \textbf{5} & 68.7 | \textbf{3} & 77.9 | \textbf{1} & 62.8 | \textbf{8} & 64.5 | \textbf{7} \\
begin & 84.6 | \textbf{5} & 84.1 | \textbf{6} & 64.4 | \textbf{11} & 81.6 | \textbf{9} & 82.0 | \textbf{8} & 86.3 | \textbf{4} & 87.1 | \textbf{2} & 79.7 | \textbf{10} & 82.6 | \textbf{7} & 87.9 | \textbf{1} & 86.4 | \textbf{3} \\
q2 & 64.3 | \textbf{10} & 72.2 | \textbf{6} & 63.7 | \textbf{11} & 77.5 | \textbf{2} & 77.4 | \textbf{3} & 64.9 | \textbf{8} & 64.8 | \textbf{9} & 80.9 | \textbf{1} & 72.7 | \textbf{4} & 70.0 | \textbf{7} & 72.4 | \textbf{5} \\
paws & 77.3 | \textbf{7} & 69.2 | \textbf{9} & 64.0 | \textbf{11} & 88.2 | \textbf{2} & 88.2 | \textbf{3} & 77.5 | \textbf{5} & 69.3 | \textbf{8} & 89.7 | \textbf{1} & 86.4 | \textbf{4} & 77.5 | \textbf{6} & 68.3 | \textbf{10} \\
\midrule
mean & 66.5 | \textbf{11} & 71.4 | \textbf{7} & 66.7 | \textbf{10} & 80.0 | \textbf{4} & 81.4 | \textbf{2} & 72.2 | \textbf{5} & 70.1 | \textbf{9} & 80.7 | \textbf{3} & 81.5 | \textbf{1} & 71.4 | \textbf{8} & 71.4 | \textbf{6} \\
\bottomrule
    \end{tabular}}
    \caption{\textbf{AUC evaluation} (x100). Bold integers: rank of model on a data set.}
    \label{tab:auceval}
\end{table*}

\begin{table*}[ht!]
    \centering
    \scalebox{0.72}{
    \begin{tabular}{llllllllllllll}
data set & \texttt{BLEU} & \texttt{QuestE} & \texttt{FactCC} & \texttt{SummaCC} & \texttt{SummacZS} & \texttt{BARTSc} & \texttt{RBERTSc} & \texttt{Q2} & \texttt{ANLI} & \texttt{DBERTSc} & \texttt{BLEURT} \\
  \midrule
qags-c & 51.9 | \textbf{9} & 58.3 | \textbf{8} & 51.9 | \textbf{10} & 59.6 | \textbf{7} & 65.5 | \textbf{4} & 68.9 | \textbf{1} & 68.5 | \textbf{2} & 63.8 | \textbf{5} & 67.7 | \textbf{3} & 60.0 | \textbf{6} & 51.9 | \textbf{11} \\
summeval & 18.5 | \textbf{9} & 51.8 | \textbf{5} & 18.4 | \textbf{10} & 82.9 | \textbf{3} & 82.6 | \textbf{4} & 22.0 | \textbf{6} & 19.1 | \textbf{8} & 86.2 | \textbf{1} & 85.5 | \textbf{2} & 21.3 | \textbf{7} & 18.4 | \textbf{11} \\
frank & 66.8 | \textbf{9} & 73.3 | \textbf{7} & 66.8 | \textbf{10} & 79.4 | \textbf{1} & 74.1 | \textbf{6} & 68.6 | \textbf{8} & 77.3 | \textbf{4} & 78.8 | \textbf{2} & 74.7 | \textbf{5} & 77.5 | \textbf{3} & 66.8 | \textbf{11} \\
qags-x & 51.5 | \textbf{7} & 51.5 | \textbf{8} & 51.5 | \textbf{9} & 61.1 | \textbf{4} & 69.0 | \textbf{2} & 52.3 | \textbf{6} & 53.1 | \textbf{5} & 62.3 | \textbf{3} & 75.7 | \textbf{1} & 51.0 | \textbf{11} & 51.5 | \textbf{10} \\
dialfact & 62.0 | \textbf{6} & 66.1 | \textbf{5} & 56.2 | \textbf{11} & 66.3 | \textbf{4} & 69.3 | \textbf{3} & 61.4 | \textbf{9} & 61.0 | \textbf{10} & 74.4 | \textbf{1} & 70.4 | \textbf{2} & 61.5 | \textbf{8} & 61.7 | \textbf{7} \\
mnbm & 89.8 | \textbf{2} & 89.8 | \textbf{3} & 88.2 | \textbf{7} & 89.4 | \textbf{5} & 88.8 | \textbf{6} & 89.8 | \textbf{1} & 87.9 | \textbf{8} & 86.7 | \textbf{9} & 73.6 | \textbf{10} & 64.2 | \textbf{11} & 89.8 | \textbf{4} \\
begin & 74.5 | \textbf{8} & 76.2 | \textbf{6} & 70.8 | \textbf{11} & 76.9 | \textbf{5} & 80.3 | \textbf{1} & 72.2 | \textbf{10} & 79.9 | \textbf{2} & 76.0 | \textbf{7} & 79.1 | \textbf{3} & 78.9 | \textbf{4} & 72.5 | \textbf{9} \\
q2 & 44.0 | \textbf{9} & 53.1 | \textbf{5} & 42.3 | \textbf{11} & 59.6 | \textbf{4} & 63.9 | \textbf{2} & 42.4 | \textbf{10} & 45.6 | \textbf{7} & 73.2 | \textbf{1} & 60.5 | \textbf{3} & 48.3 | \textbf{6} & 44.1 | \textbf{8} \\
paws & 50.6 | \textbf{8} & 49.9 | \textbf{9} & 53.0 | \textbf{7} & 80.7 | \textbf{1} & 69.2 | \textbf{4} & 46.4 | \textbf{10} & 53.7 | \textbf{6} & 73.9 | \textbf{3} & 78.9 | \textbf{2} & 67.4 | \textbf{5} & 44.4 | \textbf{11} \\
 \midrule
mean & 56.6 | \textbf{9} & 63.3 | \textbf{5} & 55.4 | \textbf{11} & 72.9 | \textbf{4} & 73.6 | \textbf{3} & 58.2 | \textbf{8} & 60.7 | \textbf{6} & 75.0 | \textbf{1} & 74.0 | \textbf{2} & 58.9 | \textbf{7} & 55.7 | \textbf{10} \\
\bottomrule
    \end{tabular}}
    \caption{\textbf{Expected accuracy evaluation} (x100). Bold integers: rank of model on a data set.}
    \label{tab:acc_eval}
\end{table*}

Given are datasets $d_1, ..., d_n$ and a diverse model $m$ that outputs a real number (`score'). It is intuitive  to transform the score into a binary prediction by fitting a logistic curve with a bias $\beta^m_0$ and a weight $\beta^m_1$, also known as \textit{Platt scaling} \cite{platt1999probabilistic}:
\begin{equation}
    \label{eq:calib}
    p(x, m)=\frac{1}{1+e^{-(\beta^m_0+ \beta^m_1 m(x))}}
\end{equation}
With this, we can make a decision with natural probability threshold $\theta=0.5$:
\begin{equation}
    f(x, m) = \begin{cases}
      1, & \text{if}\ p(x, m) > 0.5 \\
      0, & \text{otherwise}.
    \end{cases}
\end{equation}
So calibrating our model $m$ means finding suitable $\beta^m_0$, $\beta^m_1$. To calculate the generalization accuracy of $m$, it is intuitive to adopt the following strategy: For any unseen testing data set $d_i$, we calibrate Eq.\ \ref{eq:calib}, by tuning $\beta^m_0$, $\beta^m_1$ on all $d_{j \neq i}$. Finally, we get the expected accuracy on our testing data set $d_i$:
\begin{equation}
    acc(d_i) = \frac{\sum_{(x, y)\in d_i} I[f(x, m)=y]}{|d_i|}.
\end{equation}
Note that in contrast to AUC, our expected accuracy measurement is real-world oriented: Assume we have a metric such as BERTScore \cite{bertscore} -- how would an applicant transform this metric into a faithfulness predictor for filtering their generation system output? Clearly, they would need to perform calibration using development data. With our setup, we simulate this important scenario and obtain an expected accuracy score.

\section{AUC mispredicts accuracy}

\subsection{Experiment goal}

The main goal of our experiment is to investigate our hypothesis that AUC can yield a wrong picture about actual performance of models. To this aim, we conduct a real-world oriented downstream task simulation of diverse faithfulness models, measuring their expected accuracy (as detailed above).

\subsection{Experiment results}

We compare Table \ref{tab:auceval} (AUC of models) against Table \ref{tab:acc_eval} (expected accuracy). Interestingly, changes are more drastic than we had initially suspected. In fact, they even result in a \textbf{change of the best system on the benchmark}: the Q/A based system \texttt{Q2} ranks third after \texttt{ANLI} and \texttt{SummacZS} in average AUC, but according to the average accuracy, it obtains rank 1 (an improvement of two ranks). Then we also observe \textbf{interesting cases of ranking changes} of other metrics: for instance, \texttt{BLEU} yields a low rank according to AUC in the mnbm data set (rank 11), but performs much better accuracy-wise (rank 2). 

\subsection{Studying score distribution}

We saw that AUC may not predict estimated downstream accuracy. But why would some models be more negatively/positively affected by calibration? A reason may lie in the models' score distribution and their suitability for calibration. Therefore, we investigate the models' empirical distributions.

\begin{figure}
    \centering
    \includegraphics[width=0.95\linewidth]{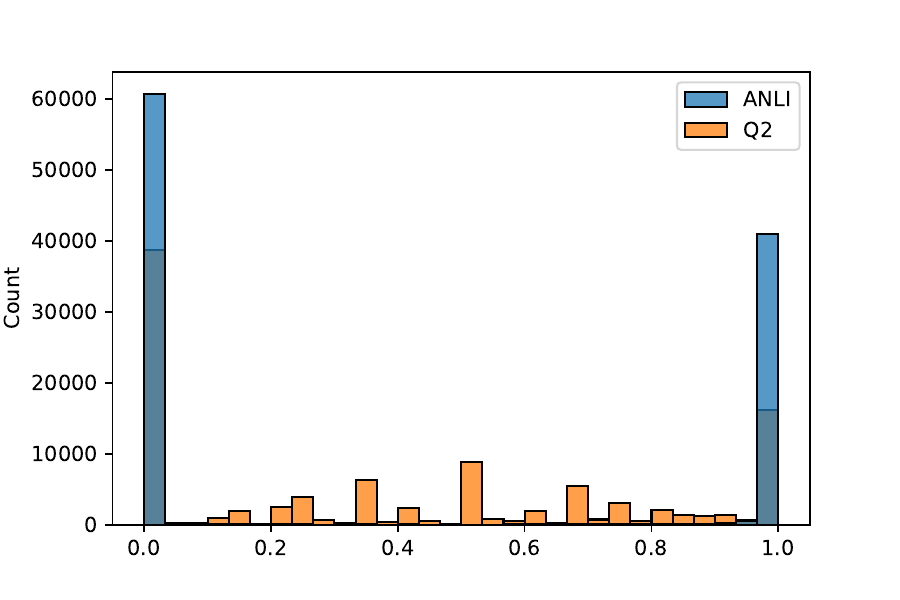}
    \caption{Histograms of best performing models \texttt{Q2} and \texttt{ANLI}. \texttt{Q2} performs best according to expected accuracy, \texttt{ANLI} performs best according to AUC.} 
    \label{fig:q2-anli-hist}
\end{figure}

\paragraph{Why would \texttt{Q2} be preferable over \texttt{ANLI}?} This question is interesting, since we saw that the best performing models differ between AUC and expected downstream accuracy. The two models are also diverse, since \texttt{Q2} employs a Q/A module while \texttt{ANLI} is an LLM trained on NLI. Their histograms (Figure \ref{fig:q2-anli-hist}) differ much: while both \texttt{ANLI} and \texttt{Q2} tend to the extremes of the spectrum, the effect is much more pronounced for \texttt{ANLI}. Throughout the scale, \texttt{Q2} appears to be more `balanced'. For the \texttt{ANLI} distribution, the data already seems harshly discriminated in two classes, perhaps increasing the difficulty of finding a generalizable threshold.

\begin{table}
    \centering
    \scalebox{0.8}{
    \begin{tabular}{l|rrrrrr|r}
          \midrule
          & \multicolumn{6}{c|}{metric names and variance} & mean \\
        better & B & Q & R & Q2 & D & &  \\
         metrics & 0.03 & 0.02 & 0.02 & 0.14 & 0.02 & & 0.05   \\ 
         \midrule
        worse  & F & SC & SZ & BA & A & BL \\
        metrics & 0.16 & 0.08 & 0.20 & 0.01 & 0.24 & 0.03 & 0.12  \\
        \bottomrule
    \end{tabular}}
    \caption{Variance of metric scores that perform better/worse under expected accuracy}.
    \label{tab:var}
\end{table}

\begin{figure}
    \centering
    \includegraphics[width=0.95\linewidth]{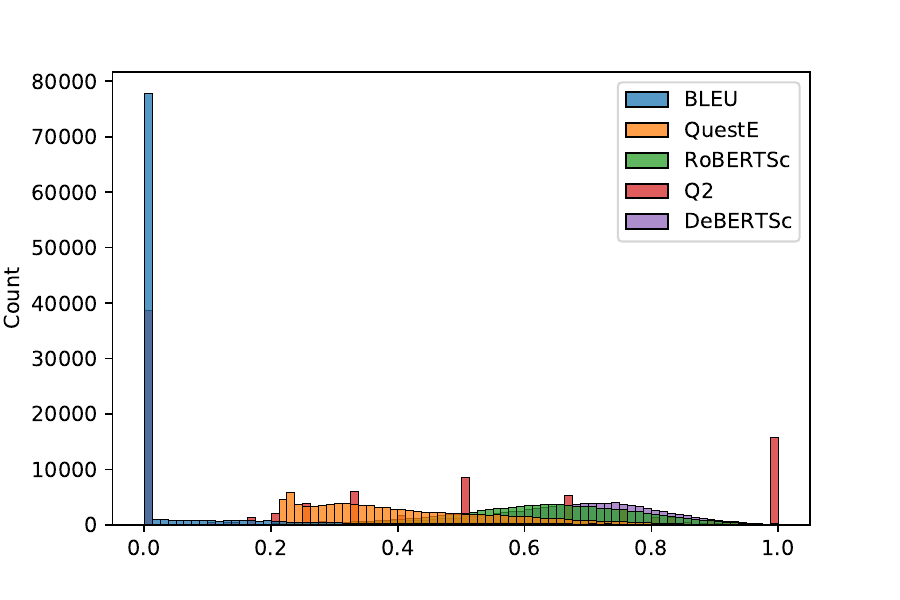}
    \includegraphics[width=0.95\linewidth]{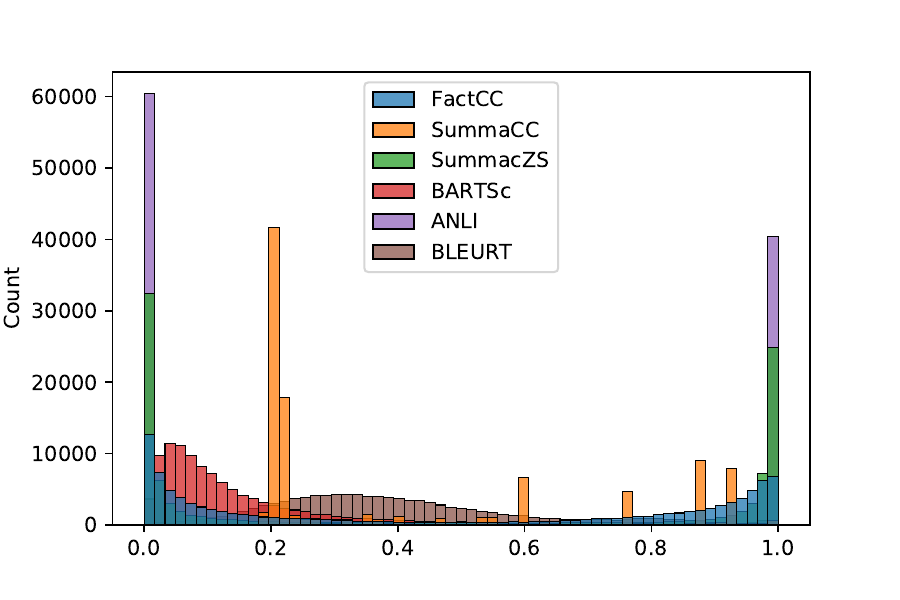}
    \caption{\textbf{Top}: histograms of models that perform \textit{better} under expected accuracy (vs.\ AUC). Bottom: histograms of models that perform \textit{worse}.}
    \label{fig:worse-better-hist}
\end{figure}

\paragraph{Less variance $\rightarrow$ easier calibration?} We create two groups of models: those that obtain a better rank according to accuracy, and those with a worse rank. The histograms are shown in Figure \ref{fig:worse-better-hist}. We see that models that are relatively more negatively affected by calibration tend to show more skewed distributions. The scores of the better models seem more balanced and exhibit a smaller average variance (Table \ref{tab:var}: avg.\ variance of better metrics=0.05; avg.\ variance of worse metrics=0.12). 

\begin{table*}[ht!]
    \centering
    \scalebox{0.69}{
    \begin{tabular}{llllllllllll|l}
metric & \texttt{BLEU} & \texttt{QuestE} & \texttt{FactCC} & \texttt{SummaCC} & \texttt{SummacZS} & \texttt{BARTSc} & \texttt{RBERTSc} & \texttt{Q2} & \texttt{ANLI} & \texttt{DBERTSc} & \texttt{BLEURT} & AVG\\
  \midrule
AUC & 66.5 | \textbf{11} & 71.4 | \textbf{7} & 66.7 | \textbf{10} & 80.0 | \textbf{4} & 81.4 | \textbf{2} & 72.2 | \textbf{5} & 70.1 | \textbf{9} & 80.7 | \textbf{3} & 81.5 | \textbf{1} & 71.4 | \textbf{8} & 71.4 | \textbf{6} & 73.9 \\
\midrule
Xdomain &  56.6 | \textbf{9} & 63.3 | \textbf{5} & 55.4 | \textbf{11} & \textbf{72.9} | \textbf{4} & \textbf{73.6} | \textbf{3} & \textbf{58.2} | \textbf{8} & \textbf{60.7} | \textbf{6} & 75.0 | \textbf{1} & 74.0 | \textbf{2} & 58.9 | \textbf{7} & 55.7 | \textbf{10} & 64.0\\
-Isotonic & 55.5 | \textbf{10} & 65.0 | \textbf{5} & \textbf{62.7} | \textbf{6} & 70.2 | \textbf{4} & 71.4 | \textbf{3} & 57.2 | \textbf{9} & 58.8 | \textbf{8} & \textbf{75.7} | \textbf{1} & 74.0 | \textbf{2} & \textbf{59.2} | \textbf{7} & 55.0 | \textbf{11} & 64.1\\
-stump & \textbf{57.6} | \textbf{9} & \textbf{68.0} | \textbf{5} & 59.3 | \textbf{6} & 70.9 | \textbf{3} & 70.6 | \textbf{4} & 56.8 | \textbf{11} & 57.6 | \textbf{8} & \textbf{75.7} | \textbf{1} & \textbf{74.1} | \textbf{2} & 58.0 | \textbf{7} & \textbf{56.9} | \textbf{10} & 64.1 \\
\midrule
OutDomain & \textbf{56.3} | \textbf{10} & 62.8 | \textbf{5} & 56.7 | \textbf{9} & \textbf{72.3} | \textbf{4} & \textbf{73.3} | \textbf{3} & \textbf{58.4} | \textbf{8} & \textbf{60.5} | \textbf{7} & 74.3 | \textbf{1} & \textbf{73.8} | \textbf{2} & \textbf{60.7} | \textbf{6} & \textbf{55.1} | \textbf{11} & 64.0 \\
-Isotonic & 55.5 | \textbf{10} & 63.0 | \textbf{5} & 60.7 | \textbf{6} & 63.8 | \textbf{4} & 71.9 | \textbf{2} & 57.2 | \textbf{9} & 58.3 | \textbf{8} & 74.7 | \textbf{1} & 71.6 | \textbf{3} & 59.6 | \textbf{7} & 55.0 | \textbf{11} & 62.8 \\
-stump&  55.4 | \textbf{10} & \textbf{63.3} | \textbf{5} & \textbf{62.7} | \textbf{6} & 64.1 | \textbf{4} & 69.1 | \textbf{3} & 56.8 | \textbf{9} & 57.6 | \textbf{8} & \textbf{75.2} | \textbf{1} & 71.7 | \textbf{2} & 58.1 | \textbf{7} & 54.7 | \textbf{11} & 62.6 \\
\midrule
OutData & 54.3 | \textbf{11} & 62.8 | \textbf{5} & 54.8 | \textbf{10} & 73.9 | \textbf{2} & 73.9 | \textbf{3} & 55.7 | \textbf{8} & 56.1 | \textbf{7} & 74.7 | \textbf{1} & 72.5 | \textbf{4} & 56.4 | \textbf{6} & 55.5 | \textbf{9} & 62.8 \\
-Isotonic& 57.9 | \textbf{6} & 61.4 | \textbf{5} & 55.5 | \textbf{9} & 67.7 | \textbf{4} & 71.4 | \textbf{2} & 55.3 | \textbf{10} & 56.2 | \textbf{8} & 73.9 | \textbf{1} & 69.0 | \textbf{3} & 54.4 | \textbf{11} & 56.8 | \textbf{7} & 61.7 \\
-stump& 56.2 | \textbf{7} & 60.7 | \textbf{5} & 55.4 | \textbf{10} & 65.2 | \textbf{4} & 67.2 | \textbf{2} & 54.6 | \textbf{11} & 55.7 | \textbf{9} & 73.5 | \textbf{1} & 67.0 | \textbf{3} & 56.0 | \textbf{8} & 56.6 | \textbf{6} & 60.7 \\
\midrule
InDomain*** &  51.2 | \textbf{10} & 69.7 | \textbf{4} & 57.3 | \textbf{8} & 74.9 | \textbf{3} & 75.7 | \textbf{2} & 61.8 | \textbf{7} & \textbf{50.8} | \textbf{11} & 77.0 | \textbf{1} & 66.9 | \textbf{5} & \textbf{51.9} | \textbf{9} & 64.2 | \textbf{6} & 63.8\\
-Isotonic*** & \textbf{65.1} | \textbf{8} & \textbf{70.6} | \textbf{5} & 61.4 | \textbf{9} & 75.3 | \textbf{4} & 77.0 | \textbf{2} & 65.4 | \textbf{7} & 48.8 | \textbf{11} & 77.3 | \textbf{1} & 76.5 | \textbf{3} & 51.0 | \textbf{10} & 67.0 | \textbf{6} & 66.9\\
-stump*** & 61.2 | \textbf{9} & 70.4 | \textbf{5} & \textbf{61.9} | \textbf{8} & \textbf{75.6} | \textbf{4} & 76.8 | \textbf{3} &\textbf{66.4} | \textbf{7} & 48.4 | \textbf{10} & \textbf{77.6} | \textbf{2} & \textbf{77.7} | \textbf{1} & 47.0 | \textbf{11} & \textbf{69.3} | \textbf{6} & 66.6 \\
\midrule
InData & 66.9 | \textbf{11} & 70.1 | \textbf{8} & 69.1 | \textbf{10} & 75.5 | \textbf{4} & 77.7 | \textbf{2} & 70.2 | \textbf{7} & 69.6 | \textbf{9} & 79.3 | \textbf{1} & 75.7 | \textbf{3} & 70.8 | \textbf{5} & 70.3 | \textbf{6} & 72.3\\
-Isotonic & \textbf{69.9} | \textbf{11} & \textbf{71.3} | \textbf{7} & \textbf{70.3} | \textbf{10} & \textbf{77.6} | \textbf{4} & \textbf{78.2} | \textbf{3} & \textbf{72.7} | \textbf{5} & \textbf{70.8} | \textbf{9} & 78.4 | \textbf{1} & \textbf{78.4} | \textbf{2} & \textbf{71.0} | \textbf{8} & \textbf{71.9} | \textbf{6} & 73.7 \\
-stump & 69.6 | \textbf{9} & 70.3 | \textbf{8} & 69.4 | \textbf{11} & 76.7 | \textbf{4} & 77.4 | \textbf{3} & 71.2 | \textbf{6} & 69.4 | \textbf{10} & \textbf{78.5} | \textbf{1} & \textbf{78.4} | \textbf{2} & 70.3 | \textbf{7} & 71.3 | \textbf{5} & 73.0\\
\bottomrule
    \end{tabular}}
    \caption{Different modes of calibration, varying calibration method and training data. Mean performance over all data sets. In each group of three lines: The first line is calibration via \textit{logistic regression}, the second line is \textit{isotonic regression}, and the third is \textit{decision stump}. Note the assumptions on data availibility: \textit{Indata} requires annotated in-domain in-distribution training; \textit{InDomain} assumes availability of in-domain training data; \textit{XDomain} assumes training data from different domains (including the testing domain), \textit{OutDomain} assumes the testing data set is from a different domain than the training data, \textit{OutData} assesses the utility of a data set to serve as calibration/training data set. ***: A data subset (PAWS) is calibrated InData (instead of Indomain), since it is the only data set of domain \textit{paraphrase/wiki}.}
    \label{tab:calib}
\end{table*}
\begin{table*}[ht!]
    \centering
    \scalebox{0.675}{
    \begin{tabular}{llllllllllll|l}
metric & \texttt{BLEU} & \texttt{QuestE} & \texttt{FactCC} & \texttt{SummaCC} & \texttt{SummacZS} & \texttt{BARTSc} & \texttt{RBERTSc} & \texttt{Q2} & \texttt{ANLI} & \texttt{DBERTSc} & \texttt{BLEURT} & AVG\\
  \midrule
AUC & 66.5 | \textbf{11} & 71.4 | \textbf{7} & 66.7 | \textbf{10} & 80.0 | \textbf{4} & 81.4 | \textbf{2} & 72.2 | \textbf{5} & 70.1 | \textbf{9} & 80.7 | \textbf{3} & 81.5 | \textbf{1} & 71.4 | \textbf{8} & 71.4 | \textbf{6} & 73.9 \\
\midrule
$\kappa$, XDomain & 12.7 | \textbf{10} & 23.1 | \textbf{5} & 14.7 | \textbf{9} & 31.4 | \textbf{4} & 34.5 | \textbf{3} & ~~8.8 | \textbf{11} & 18.9 | \textbf{7} & 40.7 | \textbf{1} & 40.7 | \textbf{2} & 19.3 | \textbf{6} & 18.9 | \textbf{8} & 24.0\\
$\kappa$, OutDomain & ~~4.3 | \textbf{10} & 17.7 | \textbf{6} & 13.7 | \textbf{8} & 30.1 | \textbf{4} & 33.9 | \textbf{3} & ~~8.5 | \textbf{9} & 17.1 | \textbf{7} & 39.8 | \textbf{1} & 39.6 | \textbf{2} & 19.3 | \textbf{5} & ~~1.4 | \textbf{11} & 20.5 \\
$\kappa$, Outdata & 14.7 | \textbf{7} & 23.2 | \textbf{5} & 10.7 | \textbf{10} & 42.8 | \textbf{3} & 45.1 | \textbf{2} & 10.5 | \textbf{8} &11.3 | \textbf{9} & 46.4 | \textbf{1} & 41.1 | \textbf{4} & ~~9.8 | \textbf{11} & 15.0 | \textbf{6} & 24.6 \\
$\kappa$, Indomain & 21.4 | \textbf{8} & 28.3 | \textbf{6} & 18.1 | \textbf{10} & 37.9 | \textbf{4} & 42.7 | \textbf{3} & 25.1 | \textbf{7} & 18.2 | \textbf{9} & 45.1 | \textbf{2} & 46.8 | \textbf{1} & 14.9 | \textbf{11} & 29.7 | \textbf{5} & 29.8 \\
$\kappa$, Indata & 21.5 | \textbf{10} & 25.2 | \textbf{8} & 17.3 | \textbf{11} & 39.9 | \textbf{4} & 41.4 | \textbf{3} & 28.9 | \textbf{5} & 24.3 | \textbf{9} & 45.1 | \textbf{1} & 42.1 | \textbf{2} & 27.1 | \textbf{6} & 26.0 | \textbf{7} & 30.8\\
\bottomrule
    \end{tabular}}
    \caption{Evaluation with KAPPA ($\kappa$) after calibration reveals the hardness of predicting faithfulness. For each model/column and calibration data/row, the best score over three calibration methods is displayed.}
    \label{tab:kappa}
\end{table*}

\section{Analysis}

\subsection{Effect of calibration technique}

We want to study the effects of different approaches to calibration. The diversity of models and data in TRUE provides an interesting study environment. Our first setup is aimed at testing the classification performance in dependence of the nature of the training data. This lets us assess domain effects and generalization power as calibration effects. For the second setup we investigate different calibration algorithms, to shed more light on the question: How to best transform a diverse model into a faithfulness assessment?

\paragraph{Setup I: Domain Effects \& Generalization.} We denote the cross-domain setup from the section before as \textit{Xdomain}. Additionally, we introduce the arguably harder setup of \textit{OutDomain} that tests the transfer to new domains: e.g., if a testing data set is from domain `summarization' we calibrate the model on `dialog' and `paraphrase' data sets. Other setups are \textit{InDomain} that allows calibration only on in-domain data. For \textit{InData} we estimate performance on a random 80/20 train/test split of a data set, averaged over 100 repetitions. Finally, in \textit{OutData} only the data set itself is used to calibrate a model, and generalization is measured on all other data sets (OutData measures the utility of a data set to serve as a development set for calibration).

\paragraph{Setup II: Calibration method effects.}

The intuitive logistic curve calibration is by far not the only possible calibration method. In fact, it has also been criticized \cite{calibrate2}, e.g., due to observed over-confidence effects. 

To test another method of probabilistic calibration, we run experiments with Isotonic regression \cite{calibrate1}. To every training datum $(x_i, y_i)$, \textit{isotonic} finds a $\hat{y}_i$ s.t.\ $(y_i - \hat{y}_i)^2$ is minimized and  $\forall j: x_j \geq x_i \implies \hat{y}_j \geq \hat{y_i}$. Prediction of an unseen datum $x_k$ is then performed through interpolation: $\hat{y}_k = \hat{y}_l + \frac{x_k - x_l}{x_r - x_l}(\hat{y}_r - \hat{y}_l)$ if $x_l \leq x_k \leq x_r$, and else either $\hat{y}_l$ (if $x_k< x_l$) or $\hat{y}_r$ (if $x_k > x_r$). Additionally, we test a non-probabilistic ($\theta \neq 0.5$) method of \textit{decision stump} that is a decision tree with depth=1, searching for one threshold that empirically best divides the training data.

\paragraph{Results.}

Table \ref{tab:calib} shows the mean (over each data set) accuracy results for variations of calibration method and variations of calibration data. We make some observations: i) As expected, \textit{InData} is the easiest setup, yielding highest accuracy (up to 73.7 accuracy with isotonic calibration). ii) Out-domain generalized calibration is hard. Here Logistic calibration provides overall best calibration (64.0 accuracy). iii) Again, there is no ranking that is same as under AUC, and all calibrated accuracy scores tend to be much lower thatn AUC. iv) different calibration methods can yield different results, but we cannot make generalizing statement as to which calibration method would be overall preferable.

Notably, only in the easy and strongly data-dependent setup of \textit{InData} calibration, AUC somewhat aligns with the expected accuracy. The relatively high scores and easiness of this setup suggest that AUC is an \textit{optimistic} performance measure, especially when data and models are diverse. 

\subsection{Other classification metrics} 

Calibrated classifiers can be evaluated with metrics other than accuracy. We show the KAPPA score as a chance-corrected accuracy measure with a random baseline score of 0.0, correcting for label skew \cite{opitz-tacl}. Results in Table \ref{tab:kappa} reveal the hardness of the task: Many measures are not much better than the chance baseline, even the best observed KAPPA score still seem low.

\section{Conclusions}

When evaluating diverse models as binary classifiers, it seems appealing to use the AUC score for benchmarking and evaluation (specifically since it factors out calibration). But we show that AUC may fail to predict the accuracy that can be expected in an application. Our work can be both interpreted as a warning to not rely (only) on AUC for evaluation as well as a call for reflecting on application when evaluating diverse \textit{decision} models.

\section{Acknowledgments}

We thank our anonymous reviewers B, C and D for their constructive comments that have improved this paper, and the LREC-COLING PC chairs for a helpful communication.

\section{Bibliographical References}\label{sec:reference}

\bibliographystyle{lrec-coling2024-natbib}
\bibliography{lrec-coling2024-example,custom}

\end{document}